\documentclass[journal,twoside,web]{ieeecolor}
\usepackage{generic}
\usepackage{amsmath,amsfonts}
\usepackage{algorithm}
\usepackage{algpseudocode}
\usepackage{multirow}
\usepackage{multicol}
\usepackage{pifont}
\usepackage{tabularx}
\usepackage{array}
\usepackage{textcomp}
\usepackage{stfloats}
\usepackage{color}
\usepackage{url}
\usepackage{verbatim}
\usepackage{graphicx}
\usepackage{cite}
\usepackage{graphicx}
\usepackage{booktabs}
\usepackage{pifont}
\usepackage{xspace}
\usepackage{bm}
\usepackage{amssymb}
\newcommand{\cmark}{\ding{52}\xspace}
\newcommand{\xmark}{\ding{56}\xspace}
\usepackage[table,xcdraw]{xcolor}

\makeatletter
\let\NAT@parse\undefined
\makeatother
\usepackage{hyperref}
\usepackage{silence}
\newcommand{\shortvec}[1]{\vec{\hspace{0.25mm}#1}}

\begin{document}

\title{Center-aware Residual Anomaly Synthesis for Multi-class Industrial Anomaly Detection}

\author{
Qiyu Chen, Huiyuan Luo, Haiming Yao, Wei Luo, Zhen Qu, \\
Chengkan Lv,~\IEEEmembership{Member,~IEEE}, Zhengtao Zhang,~\IEEEmembership{Member,~IEEE}
\thanks{Manuscript received XX XX, 2025; revised XX XX, 2025; accepted XX XX, 2025.
This work was supported in part by the National Natural Science Foundation of China under Grant No. 62303458 and 62303461.
\textit{(Corresponding author: Chengkan Lv.)} \\
\indent Qiyu Chen, Huiyuan Luo, Zhen Qu, Chengkan Lv, and Zhengtao Zhang are
with the Institute of Automation, Chinese Academy of Sciences, Beijing 100190, China,
and also with the School of Artificial Intelligence, University of Chinese Academy of Sciences, Beijing 100049, China.
(e-mail: chenqiyu2021@ia.ac.cn, huiyuan.luo@ia.ac.cn, quzhen2022@ia.ac.cn, chengkan.lv@ia.ac.cn, zhengtao.zhang@ia.ac.cn). \\
\indent Haiming Yao and Wei Luo are with the State Key Laboratory of Precision Measurement Technology and Instruments,
Department of Precision Instrument, Tsinghua University, Beijing 100084, China. (e-mail: yhm22@mails.tsinghua.edu.cn, luow23@mails.tsinghua.edu.cn).}}

\maketitle

\begin{abstract}
    Anomaly detection plays a vital role in the inspection of industrial images.
    Most existing methods require separate models for each category, resulting in multiplied deployment costs.
    This highlights the challenge of developing a unified model for multi-class anomaly detection.
    However, the significant increase in inter-class interference leads to severe missed detections.
    Furthermore, the intra-class overlap between normal and abnormal samples,
    particularly in synthesis-based methods, cannot be ignored and may lead to over-detection.
    To tackle these issues, we propose a novel Center-aware Residual Anomaly Synthesis (CRAS) method for multi-class anomaly detection.
    CRAS leverages center-aware residual learning to couple samples from different categories into a unified center,
    mitigating the effects of inter-class interference.
    To further reduce intra-class overlap,
    CRAS introduces distance-guided anomaly synthesis that adaptively adjusts noise variance based on normal data distribution.
    Experimental results on diverse datasets and real-world industrial applications demonstrate the superior detection accuracy and competitive inference speed of CRAS.
    The source code and the newly constructed dataset are publicly available at \url{https://github.com/cqylunlun/CRAS}.
\end{abstract}

\section{Introduction}
\label{sec:intro}

\IEEEPARstart{I}{ndustrial} Anomaly Detection (IAD) identifies unseen defects that deviate from the distribution of normal samples.
It has emerged as a critical component in smart manufacturing applications,
such as fabric defect detection \cite{yang2020anomaly},
plastic part inspection \cite{cao2023collaborative},
and button surface examination \cite{yao2024prior}.
Due to the scarcity of abnormal samples, current research primarily focuses on these unsupervised anomaly detection approaches.

\begin{figure*}[t]
    \centering
    \includegraphics[width=0.99\linewidth]{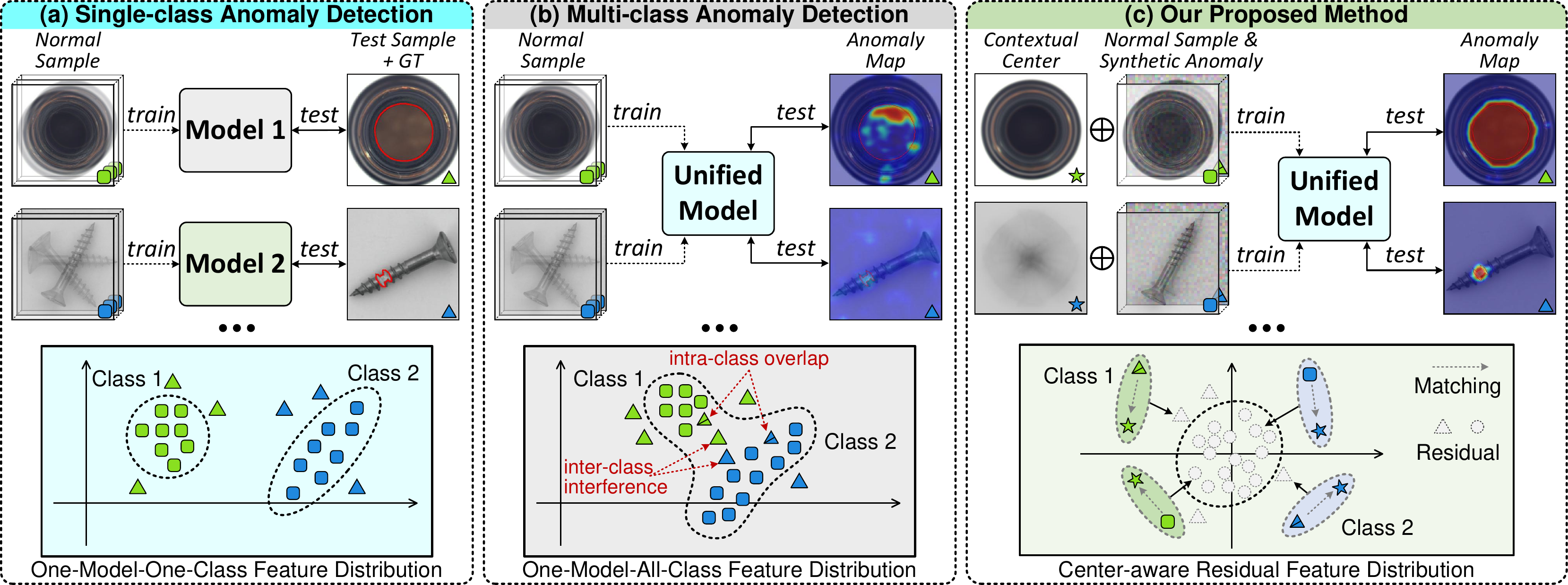}
    \caption{Conceptual illustration of Industrial Anomaly Detection (IAD) settings and our motivation.
    (a) The single-class setting trains separate models for each category independently.
    (b) The multi-class setting trains a unified model for all known categories.
    (c) Our method aims to address inter-class interference and intra-class overlap in the multi-class setting
    through center-aware residual learning and distance-guided anomaly synthesis.}
    \label{fig:concept}
\end{figure*}

As shown in the upper part of Fig.~\ref{fig:concept}(a), when detecting multiple industrial parts simultaneously,
existing IAD methods typically adopt an independent train-test paradigm for normal samples.
Each category requires an independently trained model, increasing inference complexity and memory overhead.
While the single-class setting achieves good localization, its memory usage scales poorly with category count.
As shown in the upper part of Fig.~\ref{fig:concept}(b), this study explores a more efficient and general paradigm: multi-class anomaly detection.
It addresses the high memory consumption issue by training a unified model for all categories.
However, this setting faces potential accuracy degradation due to the increased complexity in learning boundaries across multiple normal data distributions.

Existing IAD methods leverage the representation power of deep neural networks to model normal feature distributions.
Reconstruction-based methods \cite{gong2019memorizing,zavrtanik2021reconstruction,deng2022anomaly,cao2025varad,luo2025exploring}
detect anomalies by analyzing reconstruction errors but heavily rely on the quality of reconstructed images, making them prone to identity mapping issues.
Embedding-based methods \cite{salehi2021multiresolution,roth2022towards,lee2022cfa,lei2023pyramidflow}
have focused on learning more distinctive embeddings and
detecting anomalies by measuring distances between input data and representations.
As illustrated in the lower part of Fig.~\ref{fig:concept}(a-b),
normal sample boundaries can be precisely defined by separate models in the single-class setting.
However, the multi-class setting suffers from inter-class interference,
where anomalies become enclosed within unified normal boundaries, leading to significant missed detections.
Therefore, these single-class IAD approaches are unsuitable for the multi-class scenario.

To mitigate overfitting caused by the lack of anomalous representations,
popular synthesis-based methods introduce discriminative information by synthesizing anomalies based on reconstruction and embedding frameworks.
Previous approaches \cite{zavrtanik2021draem,jiang2022masked,yang2023memseg} synthesize anomalies by filling artificial textures in local image regions,
but such anomalies deviate too far from the normal patterns to provide meaningful learning signals. 
Recent approaches \cite{liu2023simplenet,chen2025unified,rolih2025supersimplenet} directly add Gaussian noise to normal features globally,
generating more discriminative subtle anomalies.
However, their fixed variance across categories often causes intra-class overlap, as shown in Fig.~\ref{fig:concept}(b), leading to severe false positives.

To address the aforementioned issues, this study proposes a novel Center-aware Residual Anomaly Synthesis (CRAS) method for multi-class IAD.
Compared to existing methods that directly learn from raw features,
our key insight is that residual features tend to exhibit more consistent distributions across different categories.
As shown in the lower part of Fig.~\ref{fig:concept}(c),
we introduce center-aware residual learning to unify samples from different categories into a shared center, mitigating inter-class interference.
Furthermore, we compute contextual centers through spatial position averaging to reduce memory costs,
and employ a global-to-local patch matching strategy to accelerate center retrieval during inference.
Motivated by the relationship between the homogeneous and heterogeneous distributions of normal samples and their distances to the centers,
we propose distance-guided anomaly synthesis that adaptively adjusts Gaussian noise based on center distances to reduce intra-class overlap.

In this way, CRAS effectively enhances anomaly detection performance, as shown in the upper part of Fig.~\ref{fig:concept}(c).
Our method comprises four core components: Multi-class Contextual Memory Center (MCMC), Hierarchical Pattern Integration (HPI),
Distance-guided Anomaly Feature Synthesis (DAFS), and Center-aware Residual Discrimination (CRD).
The main contributions of CRAS can be summarized as follows:
\begin{enumerate}
    \item 
    This study proposes a novel CRAS method for multi-class IAD,
    which employs center-aware residual learning to establish a compact unified boundary,
    efficiently mitigating missed detections due to inter-class interference.
    \item
    CRAS incorporates four key designs: MCMC, HPI, DAFS, and CRD.
    These modules enable effective discrimination between distance-guided synthetic anomalies and normal samples,
    preventing boundary ambiguity caused by intra-class overlap and ensuring fast inference.
    \item
    Extensive experiments on public datasets demonstrate the superior performance of CRAS in multi-class IAD.
    It achieves image-level and pixel-level AUROC scores of 98.3\% and 98.0\% on the representative MVTec AD dataset and outperforms competing methods on VisA and MPDD.
    Additionally, we have further validated its effectiveness in real-world Industrial Textile Defect Detection (ITDD) using our newly constructed dataset.
\end{enumerate}

\section{Related work}
\label{sec:related}

Existing IAD methods can be categorized into three types:
reconstruction-based, embedding-based, and synthesis-based.

\subsection{Reconstruction-based Methods}
\label{sec:_reconstruction}

Reconstruction-based methods \cite{gong2019memorizing,zavrtanik2021reconstruction,deng2022anomaly,cao2025varad,luo2025exploring} are widely used in anomaly detection,
assuming the model accurately reconstructs normal samples but fails to reconstruct anomalies.
These methods detect anomalies by analyzing the residuals before and after reconstruction at the image or feature level.
For multi-class anomaly detection, PSA-VT \cite{yao2024scalable} employs the Masked AutoEncoder (MAE) \cite{he2022masked} for deep feature reconstruction,
while PNPT \cite{yao2024prior} further introduces a prior normality prompt to mitigate the ``identity mapping'' problem.
To further address this issue,
HVQ-Trans \cite{lu2023hierarchical} introduces a vector quantization framework to learn discrete normal patterns and amplify anomaly deviations.
DiAD \cite{he2024diffusion} utilizes a diffusion model to improve reconstruction and semantic fidelity.
However, these methods face limitations in inference speed due to the Vision Transformer (ViT) architecture.

\subsection{Embedding-based Methods}
\label{sec:_embedding}

Recently, embedding-based methods have shown superior performance.
These methods leverage pretrained networks to extract features and compress normal features into a compact space using techniques such as
knowledge distillation \cite{salehi2021multiresolution}, memory retrieval \cite{roth2022towards},
one-class classification \cite{lee2022cfa}, normalizing flow \cite{lei2023pyramidflow} and Vision-Language Model (VLM)-based approaches \cite{cao2025personalizing}.
Anomalies are detected by measuring the distance between test features and normal clusters.
For multi-class anomaly detection, 
OMAC \cite{tan2024unsupervised} memorizes several core sets of normal patch features and uses stored global features for category querying.
HGAD \cite{yao2025hierarchical} enhances the representation capability of normalizing flows using Gaussian Mixture Model (GMM) \cite{wang2024multi}.
To mitigate the issue of ``catastrophic forgetting'' in a unified framework,
IUF \cite{tang2024incremental} reduces the risk of feature conflicts through object-aware self-attention and an iterative updating strategy.
Despite achieving competitive performance, these models have only been exposed to normal samples.

\subsection{Synthesis-based Methods}
\label{sec:_synthesis}

Since models trained only on normal samples cannot learn the anomalous distribution,
synthesis-based methods use anomaly synthesis as data augmentation,
integrating it into reconstruction-based and embedding-based frameworks.
Many works \cite{yang2020anomaly,zavrtanik2021draem,jiang2022masked,yang2023memseg,cao2023collaborative}
create binary masks of various shapes and fill them with artificial textures to synthesize image-level anomalies.
To help the model learn anomalous patterns similar to normal data distribution, some methods synthesize feature-level anomalies using
Gaussian noise perturbation \cite{liu2023simplenet,chen2025unified,rolih2025supersimplenet} or
weighted normal vectors \cite{chen2022deep, zavrtanik2022dsr,chen2024progressive}.
UniAD \cite{you2022unified} is a representative method for multi-class anomaly detection.
It employs the ViT to reconstruct features perturbed by Gaussian noise.
OneNIP \cite{gao2025learning} further incorporates global information from a single normal image prompt and
replaces Gaussian noise with anomaly texture filling.
Since the distributions of raw features vary significantly across categories and lack clear boundaries \cite{wang2023probabilistic},
these methods still suffer from inter-class interference and intra-class overlap, resulting in severe missed detections and false positives.
In contrast, our CRAS method addresses these issues through center-aware residual learning and distance-guided anomaly synthesis.
\begin{figure*}[t]
    \centering
    \includegraphics[width=0.99\linewidth]{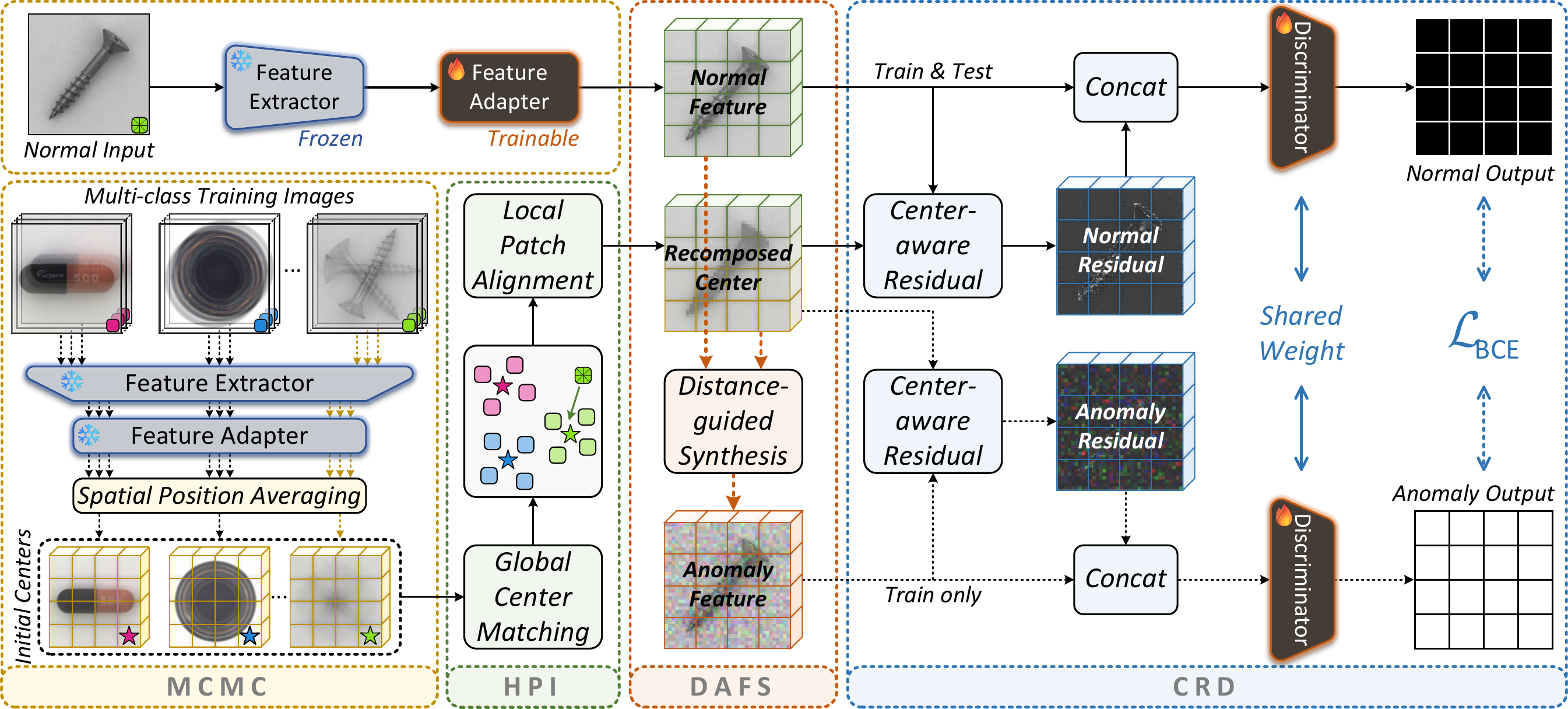}
    \caption{Schematic of the proposed CRAS.
    (a) Multi-class Contextual Memory Center (MCMC) extracts features and initializes the multi-class centers.
    (b) Hierarchical Pattern Integration (HPI) aligns normal features with the contextual centers.
    (c) Distance-guided Anomaly Feature Synthesis (DAFS) synthesizes anomaly features based on Gaussian noise.
    (d) Center-aware Residual Discrimination (CRD) enhances discriminative ability through residual learning.
    The training stage is depicted with solid and dashed arrows, while the inference stage is indicated by solid arrows.
    }
    \label{fig:schematic}
\end{figure*}

\section{Proposed Methodology: CRAS}
\label{sec:method}

\subsection{Overall Model Architecture}
\label{sec:_backbone_mlp}

The overall framework of CRAS is illustrated in Fig.~\ref{fig:schematic}.
The proposed CRAS is composed of four core modules: MCMC, HPI, DAFS, and CRD. 
The MCMC module (Section~\ref{sec:_memorized_center}) is designed to extract the normal features and initialize the multi-class contextual centers.
The HPI module (Section~\ref{sec:_pattern_integration}) then matches normal features with the class-level center and aligns corresponding vector-level features.
To synthesize more critical anomaly features while reducing intra-class overlap,
the DAFS module (Section~\ref{sec:_anomaly_synthesis}) proposes a distance-guided mechanism applied to Gaussian noise.
Finally, the CRD module (Section~\ref{sec:_residual_center}) enhances the model's discriminative ability and
mitigates inter-class interference through center-aware residual learning.
Our model is trained end-to-end using a binary classification loss.
During inference (Section~\ref{sec:_infer}),
the anomaly score is computed based on the joint discrimination of extracted test features and center-aware residuals.

\subsection{Multi-class Contextual Memory Center (MCMC)}
\label{sec:_memorized_center}

\subsubsection{Feature Extraction and Adaptation}
\label{sec:__extract}

In this paper, the MCMC module utilizes a ResNet-like backbone with frozen parameters $\phi$ pretrained on ImageNet \cite{deng2009imagenet} to extract feature representations.
Given a normal image $\bm{x}_i$ from the training set \(\mathcal{X}_\text{train}\),
where $i$ represents the sample index,
the extracted feature at hierarchy level $j$ is denoted as \mbox{$\bm{\phi}_{i,j} \in {\mathbb{R}^{{C_j} \times {H_j} \times {W_j}}}$}.
Here, $C$, $H$, and $W$ denote the number of channels, height, and width of the feature map, respectively.
To enhance robustness against minor spatial deviations,
the \mbox{$p\times p$} neighborhood patch (with $p$ denoting patch size)
of \(\vec{\phi}_{i,j}^{h,w}\) at position $(h,w)$ is aggregated using adaptive average pooling \cite{roth2022towards}.
After concatenating the locally aware features from different levels,
the extracted feature \mbox{\(\bm{t}_i={E_\phi}({\bm{x}_i})\)} is obtained, where \({{E}_{\phi}}\) represents the feature extractor.
Since industrial images typically differ from the natural images in ImageNet,
we employ a trainable MLP-like feature adapter ${{A}_{\theta}}$ to adapt the features to the target domain \cite{liu2023simplenet}.
Finally, the unbiased normal feature \(\bm{u}_i\) is obtained as \mbox{\(\bm{u}_i={{A}_{\theta}}(\bm{t}_i) \in {\mathbb{R}^{{C} \times {H} \times {W}}}\)},
where $\theta$ denotes the learnable parameters.

\begin{figure*}[t]
    \centering
    \includegraphics[width=0.99\linewidth]{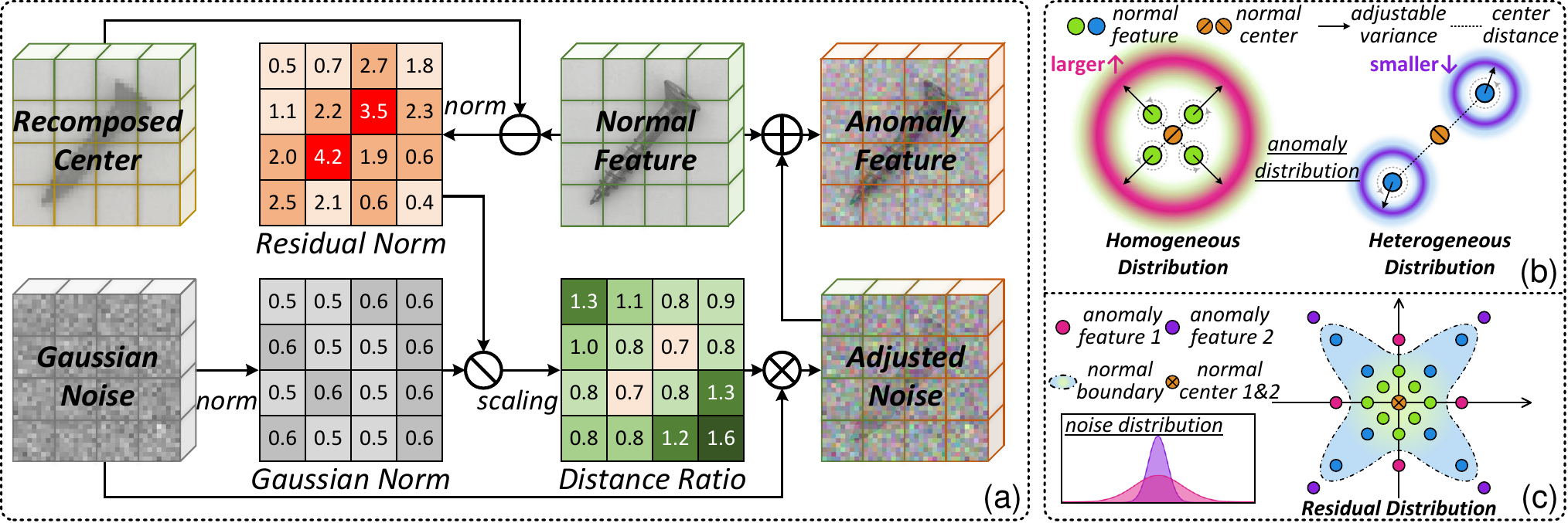}
    \caption{Overview of the DAFS module. 
    (a) Pipeline of anomaly synthesis based on recomposed center and Gaussian noise.
    (b) Anomaly feature distribution synthesized by the distance-guided mechanism.
    (c) Residual feature distribution after center-aware residual learning.
    }
    \label{fig:synthesis}
\end{figure*}

\subsubsection{Contextual Center Initialization}
\label{sec:__initial_center}

To more efficiently store the typical patterns of the multi-class training set \mbox{\(\mathcal{X}_{{\text{train}}}^k \subseteq {\mathcal{X}_{{\text{train}}}}\)},
where $k$ indicates the $k$-th class,
we introduce an initialization method based on spatial position averaging.
This method uses a pair of frozen feature extractor \({{E}_{\phi}}\) and frozen feature adapter \({{A}_{\theta}}\) to average the intra-class samples,
resulting in an initial center feature map \(\bm{c}_k\) with the same dimensions as the normal feature \(\bm{u}_i\):
\begin{equation}
    {\bm{c}_k} = \frac{1}{N_k} \sum\limits_{i=1}^{N_k} {A_\theta}({E_\phi}({\bm{x}_i})), \quad {\bm{x}_i} \in \mathcal{X}_{\text{train}}^k
    \label{eq:average}
\end{equation}
\noindent
where \(N_k\) is the number of normal samples in the \(k\)-th class.
Fig.~\ref{fig:schematic}(a) shows that \(\bm{c}_k\) preserves local patterns for classes with minimal contextual variation,
while still capturing essential information under significant variation.

\subsection{Hierarchical Pattern Integration (HPI)}
\label{sec:_pattern_integration}

Since the initial center cannot adequately handle samples with large intra-class variance,
the HPI module is proposed to obtain more appropriate recomposed centers.
As shown in Fig.~\ref{fig:schematic}(b), it principally involves two steps: global center matching and local patch alignment.
In the first step,
the feature map is flattened into a high-dimensional representation and compared against several category center vectors using cosine similarity,
ensuring efficient and accurate identification.
This coarse prediction is then refined by the second stage,
which enhances spatial consistency and mitigates the impact of subtle feature-level differences.

\subsubsection{Global Center Matching}
\label{sec:__global}

Although the multi-class labels are known during training, they are unavailable during inference.
To address this issue, we first match the input feature \(\bm{u}_i\) with the corresponding class center \(\bm{c}_k\)
by calculating their cosine similarity on the global feature map scale:
\begin{equation}
    k_i = \mathop {\arg \max }\limits_{k \in \mathcal{K}} \frac{\text{vec}(\bm{u}_i)^T \text{vec}(\bm{c}_k)}{\|\text{vec}(\bm{u}_i)\| \|\text{vec}(\bm{c}_k)\|}
    \label{eq:sim_class}
\end{equation}
\noindent
where $\mathcal{K}$ denotes the set of all class labels, and the \(\text{vec}(\cdot)\) operation flattens a matrix into a column vector.

\subsubsection{Local Patch Alignment}
\label{sec:__local}

To further refine the matching of local regions,
the matched center \(\bm{c}_{k_i}\) is used to align with the input feature \(\bm{u}_i\) at the patch level.
Given a feature vector \(\shortvec{u}_{i}^{h,w}\), we also compute the cosine similarity to search for the nearest center vector $\shortvec{c}_{{k_i}}^{h,w}$.
By iterating over all positions $(h,w)$, the recomposed center $\bm{p}_i$ is obtained as feature map:
\begin{equation}
    {\bm{p}_i} = \left\{ {\left. {\mathop {\arg \max }\limits_{\shortvec{c}_{{k_i}}^{\hat h,\hat w} \in {\bm{c}_{{k_i}}}} \frac{{\shortvec{u}_i^{h,w} \cdot
    \shortvec{c}_{{k_i}}^{\hat h,\hat w}}}{{\left\| {\shortvec{u}_i^{h,w}} \right\|\left\| {\shortvec{c}_{{k_i}}^{\hat h,\hat w}} \right\|}}} \right| \shortvec{u}_i^{h,w} \in {\bm{u}_i}} \right\}
    \label{eq:sim_patch}
\end{equation}
where $(\hat h,\hat w)$ denotes the position of the nearest center vector.

In contrast to directly performing 1-Nearest Neighbor (1-NN) search for local vectors across multi-class centers,
HPI adopts a matching mechanism that operates from global to local.
This mechanism ensures more accurate matching while significantly reducing the time complexity for a single sample from \( O(\left| \mathcal{K} \right|) \) to \( O(1) \),
where $\left| \mathcal{K} \right|$ is the number of categories.

\subsection{Distance-guided Anomaly Feature Synthesis (DAFS)}
\label{sec:_anomaly_synthesis}

Synthesizing anomalies in the feature space is an effective strategy for
improving anomaly detection tasks \cite{liu2023simplenet,you2022unified,chen2024progressive,chen2025unified}.
However, existing methods often synthesize anomalies by simply adding independent Gaussian white noise with fixed variance, lacking consideration of normal patterns.
To generate more critical anomaly features for different normal features,
we adaptively adjust the variance of Gaussian noise based on the calculated center distance.
As depicted in Fig.~\ref{fig:synthesis}(a),
the DAFS module synthesizes anomaly features by leveraging the normal feature $\bm{u}_i$, recomposed center $\bm{p}_i$,
and Gaussian noise \mbox{${\bm{g}_i} \sim \mathcal{N}(0,{\sigma ^2})$ }.
Initially, the residual norm $r_i^{h,w}$ is computed to measure the distance at each spatial position $(h,w)$ as:
\begin{equation}
    r_i^{h,w} = {\left\| {\shortvec{u}_i^{h,w} - \shortvec{p}_i^{h,w}} \right\|_2}
    \label{eq:res_norm}
\end{equation}
where \(\shortvec{u}_i^{h,w}\) is the feature vector of $\bm{u}_i$ at position $(h,w)$.

Similarly, the Gaussian norm ${g}_i^{h,w}$ is calculated from the Gaussian noise vector ${\shortvec{g}_i^{h,w}}$.
As illustrated in Fig.~\ref{fig:synthesis}(b),
for homogeneous feature distributions closer to the center,
Gaussian noise with larger variance is required to prevent intra-class overlap between anomalies and normal features.
Conversely, for heterogeneous feature distributions farther from the center,
Gaussian noise with smaller variance is essential to capture dispersed anomalous patterns accurately.

Therefore, we define the noise variance to be inversely proportional to the residual norm $r_i^{h,w}$ and directly proportional to the Gaussian norm ${g}_i^{h,w}$.
After applying linear scaling centered at the fixed point 1, the distance ratio is formulated as:
\begin{equation}
    \alpha _i^{h,w} = \beta  \cdot \left( {\frac{{g_i^{h,w}}}{{r_i^{h,w}}} \cdot
    \frac{1}{{\overline{ {{\bm{g}'_i}/{\bm{r}'_i}}}}} - 1} \right) + 1
    \label{eq:dist_ratio}
\end{equation}
where $\bm{g}'_i$ and $\bm{r}'_i$ are norm matrices, and $\beta$ is a hyperparameter that controls the magnitude of variance adjustment.
Finally, the anomaly feature $\bm{v}_i$ is synthesized by adding the adjusted Gaussian noise to the normal feature $\bm{u}_i$:
\begin{equation}
    {\bm{v}_i} = {\bm{u}_i} + {\bm{\alpha} _i} \odot {\bm{g}_i}
    \label{eq:anomaly_feat}
\end{equation}
where $\odot$ denotes the element-wise multiplication.

\subsection{Center-aware Residual Discrimination (CRD)}
\label{sec:_residual_center}

To enhance the anomaly detection capability,
we employ an MLP-like discriminator ${D_\psi}$ to conduct binary classification between normal features and synthetic anomalies,
where $\psi$ denotes the learnable parameters.
However, if the model only learns from the raw feature distribution, it struggles to detect anomalies across different classes.
To address this limitation, the CRD module is introduced to further learn the center-aware residual distribution, mitigating missed detection issues.
Specifically, center-aware features $\bm{y}_i$ and $\bm{z}_i$
are obtained by concatenating the normal and anomalous features with their corresponding center-aware residuals:
\begin{equation}
    \left[ {\begin{array}{*{20}{c}}
    {\bm y_i}\\
    {\bm z_i}
    \end{array}} \right] = \left[ {\begin{array}{*{20}{c}}
    { {\bm u_i} \mid \bm u_i - \bm p_i}\\
    { {\bm v_i} \mid \bm v_i - \bm p_i}
    \end{array}} \right]
    \label{eq:concat}
\end{equation}
where \mbox{${\bm y_i, \bm z_i} \in \mathbb{R}^{2C \times H \times W}$}.
The feature adapter ${{A}_{\theta}}$ and the discriminator ${D_\psi }$ are subsequently trained to distinguish between negative and positive features:
\begin{equation}
    {\mathcal L} = \frac{1}{N}\frac{1}{{HW}}\sum\limits_{i = 1}^N {\sum\limits_{h,w} {{f_{\text{BCE}}}\left( {{D_\psi }\left( {\left[ {\begin{array}{*{20}{c}}
    {\shortvec y_i^{h,w}}\\
    {\shortvec z_i^{h,w}}
    \end{array}} \right]} \right),\left[ {\begin{array}{*{20}{c}}
    0\\
    1
    \end{array}} \right]} \right)} }
    \label{eq:loss}
\end{equation}
where $N$ is the total number of samples across all classes, and $f_{\text{BCE}}$ denotes the binary cross-entropy loss function.

\begin{figure}[t]
    \centering
    \includegraphics[width=0.99\linewidth]{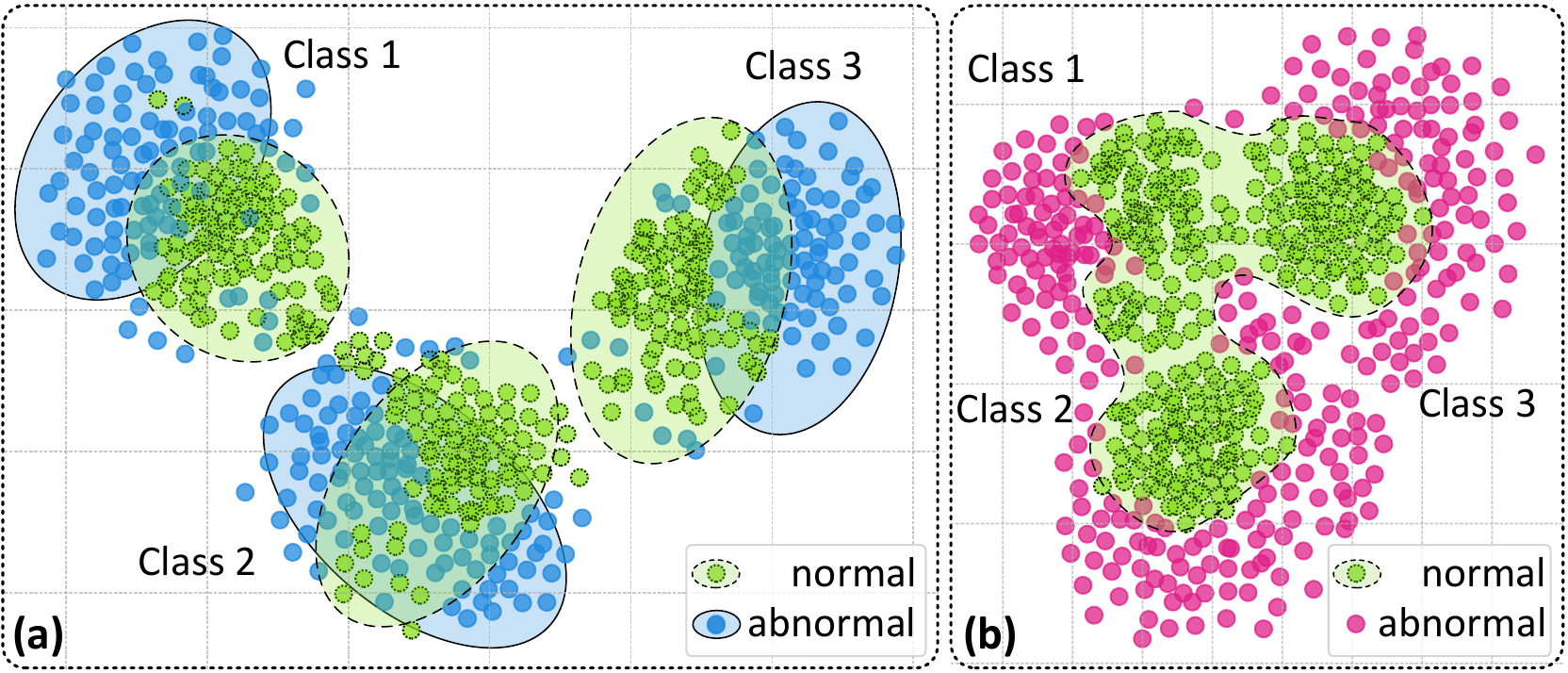}
    \caption{t-SNE visualization of adapted patch-level feature distributions.
    (a) Raw features show significant overlap between normal and abnormal samples.
    (b) Center-aware residual features generated by CRD form more compact normal clusters and better separation from anomalies.
    }
    \label{fig:tsne}
\end{figure}

As shown in Fig.~\ref{fig:synthesis}(c), features from different classes are coupled around the same center.
To further validate the interpretability of our method,
we visualize the adapted patch-level feature distributions before the discriminator using t-SNE.
As shown in Fig.~\ref{fig:tsne}(a), raw features exhibit significant overlap between normal and abnormal samples within each class.
In contrast, residual features generated by CRD form more compact and separable clusters, where anomalies are pushed farther from class centers in Fig.~\ref{fig:tsne}(b).
This suggests that center-aware residuals effectively enhance anomaly discrimination, thereby reducing false and missed detections.

Overall, CRAS improves multi-class anomaly detection performance
by addressing intra-class overlap through anomaly synthesis with DAFS and inter-class interference through residual learning with CRD.

\subsection{Anomaly Scoring at Inference Stage}
\label{sec:_infer}

As depicted in Fig.~\ref{fig:schematic}, the inference stage is indicated by the solid arrows.
Given an unseen image $\bm{x_i}$ from the test set \(\mathcal{X}_\text{test}\),
we first extract the unbiased test feature \(\bm{u}_i\) using the feature extractor \({{E}_{\phi}}\) and feature adapter \({{A}_{\theta}}\).
Next, \(\bm{u}_i\) is matched to one of the multi-class contextual centers \(\bm{c}\),
and its local patches are aligned to derive the recomposed center \(\bm{p}_i\).
Finally, the discriminator ${D_\psi }$ outputs the confidence score for the center-aware feature $\bm{y}_i$ derived from \(\bm{u}_i\) and \(\bm{p}_i\).

After bilinear interpolation $f_{{\text{resize}}}$ and Gaussian filtering $f_{{\text{smooth}}}$, we obtain the anomaly score for the segmentation task:
\begin{equation}
    {\bm{\mathcal S}_{{\text{seg}}}} = {f_{{\text{smooth}}}}\left( {f_{{\text{resize}}}^{{H_0},{W_0}}\left( {{D_\psi }\left( {\left[ {
        {{\bm{u}_i}} \mid {\bm{u}_i} - {\bm{p}_i}} \right]} \right)} \right)} \right)
    \label{eq:score_seg}
\end{equation}
where \({H_0}\) and \({W_0}\) denote the dimensions of the input image $\bm{x_i}$.
To obtain the anomaly score for the classification task, we take the maximum value across all positions:
\begin{equation}
    \mathcal{S}_{\text{cls}} = \max_{(h, w)} \mathcal{S}_{\text{seg}}^{h,w}
    \label{eq:score_cls}
\end{equation}
where \(\mathcal{S}_{\text{seg}}^{h,w}\) is the pixel-level anomaly score at position $(h,w)$.

\begin{table*}[t]\scriptsize
   \centering
   \caption{Performance comparison of various methods on MVTec AD under multi-class setting, as measured by I-AUROC\%/P-AUROC\%.
   The best results for each category are highlighted in bold, and the second-best results are underlined.}
     \begin{tabularx}{\textwidth}{p{1.2cm}|*{2}{>{\centering}X}|*{2}{>{\centering}X}|*{4}{>{\centering}X}|*{3}{>{\centering}X}|>{\centering\arraybackslash\columncolor{pink!30}}X}
     \noalign{\hrule height 0.4mm}
     \multirow{2}{*}{Taxonomy} & \multicolumn{4}{c|}{\textit{w/o} Anomaly Synthesis} & \multicolumn{8}{c}{\textit{w/} Anomaly Synthesis} \\
     \cline{2-13}      & \multicolumn{2}{c|}{Reconstruction-based} & \multicolumn{2}{c|}{Embedding-based} & \multicolumn{4}{c|}{Reconstruction-based} & \multicolumn{4}{c}{Embedding-based} \\
     \hline
     Method~$\rightarrow$ & RD4AD & PNPT  & CFA   & PFlow & DRAEM & DSR   & UniAD & OneNIP & SimpleNet & GLASS & PBAS  &  \\
     Category~$\downarrow$ & \cite{deng2022anomaly} & \cite{yao2024prior} & \cite{lee2022cfa} & \cite{lei2023pyramidflow} &
                             \cite{zavrtanik2021draem} & \cite{zavrtanik2022dsr} & \cite{you2022unified} & \cite{gao2025learning} &
                             \cite{liu2023simplenet} & \cite{chen2025unified} & \cite{chen2024progressive} & \multirow{-2}{*}{\textbf{CRAS}} \\
     \hline
     Carpet & 99.5/98.1 & 98.0/97.5 & 73.0/77.4 & 56.7/88.0 & 87.8/93.5 & 89.7/95.5 & \textbf{99.8}/\underline{98.6} & \underline{99.6}/\textbf{98.8} & 98.0/96.9 & 97.4/95.9 & 97.9/\underline{98.6} & 98.9/\underline{98.6} \\
     Grid  & 98.2/\underline{97.7} & 98.3/96.9 & 38.6/40.0 & 82.3/74.0 & 57.1/90.3 & 98.3/95.0 & 92.1/94.4 & \textbf{100}/96.8 & 65.2/51.9 & 81.0/85.5 & 66.3/81.7 & \underline{99.3}/\textbf{98.1} \\
     Leather & \textbf{100}/98.8 & \underline{98.5}/97.7 & 78.4/27.0 & 77.0/94.4 & 83.5/92.1 & 97.8/92.6 & \textbf{100}/\underline{99.2} & \textbf{100}/\textbf{99.6} & 96.6/94.6 & \textbf{100}/98.9 & \textbf{100}/99.1 & \textbf{100}/99.1 \\
     Tile  & \underline{98.9}/94.5 & 97.0/93.8 & 55.3/48.3 & 96.6/90.6 & 74.2/89.9 & 89.2/89.5 & 96.8/88.9 & 98.8/92.2 & 91.1/83.1 & 98.0/\textbf{96.4} & 98.6/93.6 & \textbf{99.8}/\underline{96.3} \\
     Wood  & 99.2/93.1 & 98.1/\underline{94.7} & 68.3/58.9 & \underline{99.6}/88.8 & 85.3/81.2 & 91.3/85.8 & 98.9/93.5 & \underline{99.6}/94.3 & \textbf{99.9}/92.1 & 99.4/\textbf{96.9} & \underline{99.6}/92.8 & \underline{99.6}/93.5 \\
     Bottle & 56.5/96.2 & 98.5/96.8 & 59.2/53.6 & \underline{99.4}/\underline{97.7} & 71.9/70.5 & 95.0/84.9 & 98.5/96.9 & 98.7/97.6 & 96.5/88.5 & 94.4/91.3 & \underline{99.4}/95.8 & \textbf{100}/\textbf{98.7} \\
     Cable & 96.5/92.9 & 96.6/95.9 & 63.4/37.4 & 70.3/92.2 & 60.9/60.7 & 74.7/65.6 & 91.8/93.3 & \textbf{98.4}/\underline{96.8} & 86.7/85.5 & 72.2/82.3 & 93.1/92.7 & \underline{96.7}/\textbf{98.2} \\
     Capsule & 90.3/96.9 & \underline{91.8}/97.0 & 60.2/57.7 & 70.1/97.4 & 51.9/57.6 & 76.7/72.8 & 72.8/97.2 & 86.4/\underline{97.9} & 84.0/91.9 & 85.9/95.1 & 83.9/95.1 & \textbf{96.2}/\textbf{98.8} \\
     Hazelnut & \textbf{100}/97.5 & 98.5/97.4 & 64.3/85.8 & 93.4/97.6 & 62.4/79.7 & 83.9/86.8 & 96.9/97.0 & \textbf{100}/\underline{97.9} & 98.4/93.4 & 98.1/96.9 & 98.9/96.4 & \underline{99.8}/\textbf{98.3} \\
     Metal nut & \textbf{100}/91.5 & 98.1/94.3 & 72.0/75.0 & 84.0/94.7 & 72.8/82.9 & 77.3/88.5 & 96.4/91.7 & 99.0/\underline{95.5} & 73.7/77.1 & 91.6/86.1 & 91.3/94.5 & \underline{99.6}/\textbf{98.6} \\
     Pill  & \underline{94.4}/94.3 & 94.2/94.2 & 68.6/69.6 & 80.9/95.6 & 38.0/86.3 & 79.8/80.3 & 70.9/87.8 & 83.9/87.9 & 84.2/93.1 & 84.7/\underline{95.9} & 80.3/92.7 & \textbf{96.3}/\textbf{98.6} \\
     Screw & 91.8/95.8 & \underline{94.4}/98.0 & 48.0/71.6 & 65.7/89.7 & 27.3/78.8 & 80.1/90.9 & 79.6/97.6 & \textbf{96.7}/\textbf{98.9} & 67.0/79.4 & 86.5/\underline{98.3} & 68.4/96.7 & 88.9/97.8 \\
     Toothbrush & \underline{98.6}/98.2 & 98.5/97.5 & 52.5/68.3 & \underline{98.6}/\underline{98.6} & 60.6/83.7 & 75.8/89.3 & 92.5/98.3 & 96.1/\underline{98.6} & 88.3/93.5 & 83.1/95.8 & 93.6/96.8 & \textbf{100}/\textbf{99.0} \\
     Transistor & 95.4/89.7 & \underline{98.2}/95.8 & 54.5/70.4 & 82.6/93.8 & 68.8/57.0 & 66.5/77.4 & 97.7/96.5 & \textbf{100}/\textbf{97.2} & 78.6/78.4 & 81.5/80.4 & 97.8/91.8 & \textbf{100}/\underline{97.1} \\
     Zipper & 97.3/94.6 & 95.9/96.8 & 43.7/62.7 & 81.9/92.4 & 87.9/94.3 & 89.9/70.0 & 82.8/96.0 & 96.2/97.0 & 96.5/94.7 & 91.5/97.4 & \underline{97.8}/\underline{98.6} & \textbf{99.7}/\textbf{98.9} \\
     \hline
     Average & 94.4/95.3 & \underline{97.0}/96.3 & 60.0/60.3 & 82.6/92.4 & 66.0/79.9 & 84.4/84.3 & 91.2/95.1 & 96.9/\underline{96.5} & 87.0/86.3 & 89.7/92.9 & 91.1/94.4 & \textbf{98.3}/\textbf{98.0} \\
     \noalign{\hrule height 0.4mm}
     \end{tabularx}%
   \label{tab:mvtec_compare}%
 \end{table*}%

\section{Experiment and Analysis}
\label{sec:exper}
This section provides the experimental evaluation of the proposed CRAS across several industrial image datasets,
including comparative experiments, ablation studies, parameter analyses, and real-world industrial application. 

\subsection{Experimental Setup}
\label{sec:_setup}

\subsubsection{Dataset Descriptions}
\label{sec:__dataset}

This study evaluates the proposed CRAS on three public industrial image datasets:
MVTec AD \cite{bergmann2019mvtec}, a challenging dataset with 15 categories of high-resolution industrial products featuring over 70 types of defects;
VisA \cite{zou2022spot}, one of the largest industrial anomaly detection datasets with 12 categories of colored industrial parts;
and MPDD \cite{jezek2021deep}, a dataset of metal parts under varied conditions across 6 categories.
Additionally, we have collected and annotated the Industrial Textile Defect Detection (ITDD) dataset from industrial site,
as detailed in Section~\ref{sec:_itdd}. 

\begin{table}[t]\footnotesize
  \centering
  \caption{Comparative results of computational costs. The best results are in bold, and the second-best are underlined.}
  \begin{tabularx}{0.42\textwidth}{>{\centering\arraybackslash}p{1cm}>{\centering\arraybackslash}p{1cm}|*{2}{>{\centering\arraybackslash}X}}
    \noalign{\hrule height 0.4mm}
    \multicolumn{2}{c|}{Metric~$\rightarrow$} & Memory Usage & Frame Rate \\
    \multicolumn{2}{c|}{Method~$\downarrow$} & (MB)    & (FPS) \\
    \hline
    RD4AD & \cite{deng2022anomaly}   & 10161 & 59 \\
    PNPT  & \cite{yao2024prior}   & 11081 & 45 \\
    UniAD & \cite{you2022unified}   & \textbf{5641} & 21 \\
    OneNIP & \cite{gao2025learning}   & 10232 & 25 \\
    GLASS & \cite{chen2025unified}   & 7680  & \textbf{80} \\
    \hline
    \rowcolor{pink!30} \multicolumn{2}{c|}{\textbf{CRAS}} & \underline{6706}  & \underline{77} \\
    \noalign{\hrule height 0.4mm}
    \end{tabularx}%
  \label{tab:fps}%
\end{table}%

\subsubsection{Implementation Details}
\label{sec:__implement}

The experiments were conducted on a computing system equipped with
Intel\textregistered~Xeon\textregistered~Silver 4210R CPUs @2.40GHz and NVIDIA GeForce RTX 4090 GPUs.
All methods were re-trained and tested under a unified multi-class anomaly detection setting to ensure fair comparisons.
Input images were resized to $329\times329$ and center-cropped to $288\times288$,
following similar preprocessing strategies used in the official implementations of
SimpleNet \cite{liu2023simplenet}, GLASS \cite{chen2025unified}, and PBAS \cite{chen2024progressive}.
To maintain consistency and account for computational constraints,
all models were trained for 100 epochs, as recommended by the ADer survey \cite{zhang2024ader} on multi-class IAD.
For the training of our method, the batch size was set to 32.
The AdamW optimizer was employed to train the feature adapter \(A_{\theta}\) and the discriminator \(D_{\psi}\)
with learning rates of \mbox{\(1\times10^{-4}\)} and \mbox{\(2\times10^{-4}\)}, respectively.
Our proposed CRAS framework comprises four modules: MCMC, HPI, DAFS, and CRD.
For MCMC, we used WideResNet50 \cite{zagoruyko2016wide} pretrained on ImageNet as the backbone for the feature extractor \(E_{\phi}\).
We then aggregated features from hierarchy levels 2 and 3 using a neighborhood patch size \(p\) of 3.
For DAFS, the variance $\sigma$ of Gaussian noise was set to 0.015 and the anomaly magnitude $\beta$ was set to 0.3.

\subsubsection{Evaluation Metrics}
\label{sec:__metrics}

To evaluate the binary classification performance at both image and pixel levels,
we use the Area Under the Receiver Operating Characteristic Curve (AUROC) and Average Precision (AP).
AUROC and AP are denoted as I-AUROC/I-AP for the image level and P-AUROC/P-AP for the pixel level.
Compared to AUROC, AP provides greater insight for imbalanced positive and negative samples.

\begin{figure}[t]
  \centering
  \includegraphics[width=0.99\linewidth]{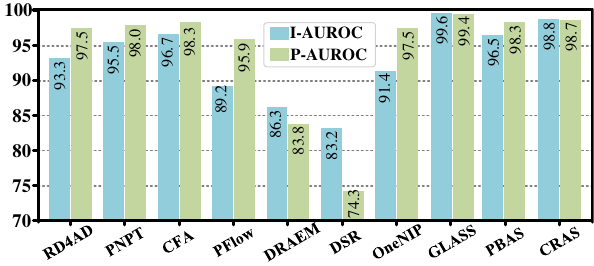}
  \caption{Performance comparison of various methods on MPDD under the single-class setting, as measured by I-AUROC\% and P-AUROC\%.}
  \label{fig:mpdd_s_compare}
\end{figure}

\subsection{Comparative Experiments}
\label{sec:_comparative}

The proposed CRAS is compared with several typical and state-of-the-art (SOTA) methods on three IAD datasets.
Reconstruction-based methods include RD4AD \cite{deng2022anomaly} and PNPT \cite{yao2024prior},
while embedding-based methods include CFA \cite{lee2022cfa} and PFlow \cite{lei2023pyramidflow}.
For methods integrating reconstruction-based anomaly synthesis, comparisons are made with DRAEM \cite{zavrtanik2021draem},
DSR \cite{zavrtanik2022dsr}, UniAD \cite{you2022unified}, and OneNIP \cite{gao2025learning}.
Similarly, embedding-based anomaly synthesis methods such as SimpleNet \cite{liu2023simplenet},
GLASS \cite{chen2025unified}, and PBAS \cite{chen2024progressive} are also evaluated.

\subsubsection{Results on MVTec AD}
\label{sec:__results_mvtec}

As shown in Table~\ref{tab:mvtec_compare}, the proposed CRAS outperforms all competitors at both image and pixel levels on the MVTec AD dataset.
Notably, CRAS almost achieves optimal or near-optimal performance across all 15 categories, demonstrating its robustness and versatility.
Compared to SimpleNet \cite{liu2023simplenet} which serves as our baseline model, CRAS exhibits improvements of +11.3\% in I-AUROC and +11.7\% in P-AUROC.
It is noteworthy that UniAD \cite{you2022unified}, PNPT \cite{yao2024prior}, and OneNIP \cite{gao2025learning} have been specifically tailored for multi-class anomaly detection,
thereby achieving relatively good performance.
Our method further surpasses the latest SOTA method PNPT, with improvements of +1.3\% in I-AUROC and +1.7\% in P-AUROC.
Moreover, Table~\ref{tab:fps} presents the training memory usage and testing frame rate of representative methods under the same batch size,
demonstrating that CRAS not only achieves superior performance but also offers competitive computational efficiency due to its lightweight architecture.

\begin{figure*}[t]
  \centering
  \includegraphics[width=0.99\linewidth]{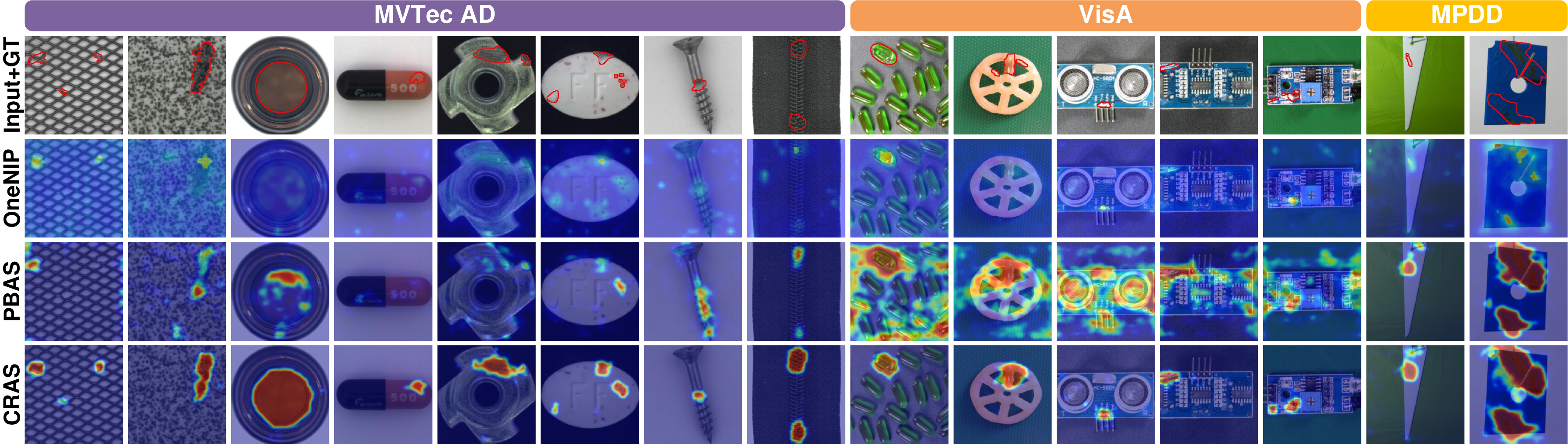}
  \caption{Qualitative comparison of CRAS with recent SOTA methods
  (OneNIP \cite{gao2025learning} and PBAS \cite{chen2024progressive}) across different categories of several datasets.}
  \label{fig:visual}
\end{figure*}

\begin{table*}[t]\scriptsize
   \centering
   \caption{Performance comparison of various methods on VisA under multi-class setting, as measured by I-AUROC\%, I-AP\%, P-AUROC\%, and P-AP\%.
   The best results are highlighted in bold, and the second-best results are underlined.}
   \begin{tabularx}{\textwidth}{p{1.2cm}|*{11}{>{\centering}X}|>{\centering\arraybackslash\columncolor{pink!30}}X}
     \noalign{\hrule height 0.4mm}
     Method~$\rightarrow$ & RD4AD & PNPT  & CFA   & PFlow & DRAEM & DSR   & UniAD & OneNIP & SimpleNet & GLASS & PBAS  &  \\
     Metric~$\downarrow$ & \cite{deng2022anomaly} & \cite{yao2024prior} & \cite{lee2022cfa} & \cite{lei2023pyramidflow} &
                           \cite{zavrtanik2021draem} & \cite{zavrtanik2022dsr} & \cite{you2022unified} & \cite{gao2025learning} &
                           \cite{liu2023simplenet} & \cite{chen2025unified} & \cite{chen2024progressive} & \multirow{-2}{*}{\textbf{CRAS}} \\
     \hline
     I-AUROC & 85.6  & 92.2  & 56.9  & 80.9  & 59.8  & 86.0  & 83.3  & \underline{92.4}  & 89.2  & 90.8  & 87.2  & \textbf{93.5} \\
     I-AP  & 87.6  & 93.9  & 65.4  & 83.5  & 65.9  & 89.7  & 85.8  & \underline{94.0}  & 92.4  & 93.3  & 90.5  & \textbf{95.2} \\
     P-AUROC & 95.9  & \underline{97.0}  & 71.2  & 93.7  & 67.1  & 92.3  & 96.3  & \textbf{97.7} & 96.6  & 94.3  & 95.8  & \textbf{97.7} \\
     P-AP  & 30.5  & 38.0  & 12.4  & 20.4  & 3.25  & 37.6  & 28.4  & \textbf{43.6} & 36.3  & 30.6  & 34.0  & \underline{38.4} \\
     \noalign{\hrule height 0.4mm}
   \end{tabularx}%
   \label{tab:visa_compare}%
 \end{table*}%

\begin{table*}[t]\scriptsize
   \centering
   \caption{Performance comparison of various methods on MPDD under multi-class setting, as measured by I-AUROC\%, I-AP\%, P-AUROC\%, and P-AP\%.
   The best results are highlighted in bold, and the second-best results are underlined.}
   \begin{tabularx}{\textwidth}{p{1.2cm}|*{11}{>{\centering}X}|>{\centering\arraybackslash\columncolor{pink!30}}X}
   \noalign{\hrule height 0.4mm}
   Method~$\rightarrow$ & RD4AD & PNPT  & CFA   & PFlow & DRAEM & DSR   & UniAD & OneNIP & SimpleNet & GLASS & PBAS  &  \\
   Metric~$\downarrow$ & \cite{deng2022anomaly} & \cite{yao2024prior} & \cite{lee2022cfa} & \cite{lei2023pyramidflow} &
                         \cite{zavrtanik2021draem} & \cite{zavrtanik2022dsr} & \cite{you2022unified} & \cite{gao2025learning} &
                         \cite{liu2023simplenet} & \cite{chen2025unified} & \cite{chen2024progressive} & \multirow{-2}{*}{\textbf{CRAS}} \\
     \hline
     I-AUROC & 91.4  & 93.6  & 90.8  & 73.8  & 66.4  & 75.7  & 61.2  & 82.4  & 91.9  & \underline{94.4}  & 93.7  & \textbf{95.0} \\
     I-AP  & 93.3  & 95.5  & 92.9  & 80.2  & 73.6  & 78.8  & 70.6  & 84.4  & 94.8  & \underline{95.8}  & 95.5  & \textbf{96.3} \\
     P-AUROC & 97.5  & 96.4  & 94.2  & 90.2  & 78.8  & 71.9  & 92.7  & 95.2  & 97.7  & 97.3  & \underline{97.8}  & \textbf{98.3} \\
     P-AP  & 38.2  & 36.1  & 34.7  & 16.0  & 26.5  & 19.3  & 14.5  & 24.0  & 37.4  & 38.1  & \underline{39.0}  & \textbf{39.2} \\
     \noalign{\hrule height 0.4mm}
     \end{tabularx}%
   \label{tab:mpdd_compare}%
 \end{table*}%

\begin{table}[t]\scriptsize
  \centering
  \caption{Performance ablation of different components in CRAS on several datasets, as measured by I-AUROC\% and P-AUROC\%.
  The best results are highlighted in bold.}
    \begin{tabularx}{0.49\textwidth}{>{\centering}p{1.3cm}|*{5}{>{\centering}X}|>{\centering\arraybackslash\columncolor{pink!30}}X}
    \noalign{\hrule height 0.4mm}
    \multirow{2}{*}{Module↓} & \multicolumn{5}{c|}{Model Variants}   &  \\
\cline{2-6}          & (A)   & (B)   & (C)   & (D)   & (E)   & \multirow{-2}{*}{\textbf{Ours}} \\
    \hline
    \textbf{MCMC} & \textcolor{gray}{\xmark}     & \textcolor{gray}{\xmark}     & \cmark     & \cmark     & \cmark     & \cmark \\
    \textbf{HPI} & \textcolor{gray}{\xmark}     & \cmark     & \textcolor{gray}{\xmark}     & \cmark     & \cmark     & \cmark \\
    \textbf{DAFS} & \textcolor{gray}{\xmark}     & \cmark     & \cmark     & \textcolor{gray}{\xmark}     & \cmark     & \cmark \\
    \textbf{CRD} & \textcolor{gray}{\xmark}     & \cmark     & \cmark     & \cmark     & \textcolor{gray}{\xmark}     & \cmark \\
    \hline
    MVTec AD & 94.9/96.8 & 97.6/97.5 & 97.8/97.8 & 97.6/97.6 & 95.8/96.9 & \textbf{98.3/98.0} \\
    VisA  & 89.5/97.5 & 90.7/96.7 & 92.4/97.7 & 92.1/97.5 & 90.6/97.2 & \textbf{93.5/97.7} \\
    MPDD  & 90.9/97.5 & 80.2/96.9 & 94.7/97.4 & 94.4/95.8 & 91.4/98.0 & \textbf{95.0/98.3} \\
    \noalign{\hrule height 0.4mm}
    \end{tabularx}%
  \label{tab:ablation}%
\end{table}%

\subsubsection{Results on VisA}
\label{sec:__results_visa}

As a large-scale and multi-part dataset, VisA presents a more challenging scenario compared to MVTec AD.
The outcomes in Table~\ref{tab:visa_compare} demonstrate that the proposed CRAS achieves the superior performance on the VisA dataset.
Compared to our baseline model SimpleNet, CRAS exhibits improvements of +4.3\% in I-AUROC and +1.1\% in P-AUROC.
It is evident that most methods with anomaly synthesis strategies outperform those without,
indicating the importance of anomaly synthesis in multi-class anomaly detection.
Furthermore, CRAS surpasses the second-best method OneNIP with +1.1\% improvement in I-AUROC.

\subsubsection{Results on MPDD}
\label{sec:__results_mpdd}

Additionally, we conducted further experiments on the MPDD dataset, which includes images taken from diverse shooting angles.
As shown in Table~\ref{tab:mpdd_compare}, the proposed CRAS achieves the best performance on the MPDD dataset.
To further validate the anomaly detection capabilities of CRAS,
we also conducted experiments on the MPDD dataset under the single-class setting.
Fig.~\ref{fig:mpdd_s_compare} demonstrates that while GLASS \cite{chen2025unified} achieves the best performance using this one-model-one-class training strategy,
CRAS remains competitive with the second-best performance.

\subsubsection{Qualitative Results}  
\label{sec:__visual_results}

As shown in Fig.~\ref{fig:visual}, our CRAS is qualitatively compared with two recent SOTA methods.
All three methods employ anomaly synthesis strategies for simulating anomalous patterns.
Due to the diverse categories in MVTec AD,
OneNIP \cite{gao2025learning} and PBAS \cite{chen2024progressive} face severe missed detections due to inter-class interference.
In contrast, our method addresses this with center residual learning for precise pixel-level segmentation.
PBAS localizes most anomalies on VisA but suffers from over-detection due to intra-class overlap in normal regions,
while our distance-guided synthesis during training ensures a clean background and strong anomaly responses.
For the dispersed industrial parts in MPDD, our method maintains accurate anomaly coverage.

\subsection{Ablation Studies}
\label{sec:_ablation}

In this section, ablation experiments were conducted to verify the effectiveness of each component in the proposed CRAS.
Table~\ref{tab:ablation} shows the ablation results of different model variants on several datasets.
Among them, model variant (A) serves as the baseline, similar to the framework of SimpleNet \cite{liu2023simplenet}.
It uses the feature extractor and the feature adapter during training to obtain normal features,
which are then combined with synthetic Gaussian anomalies and fed into the discriminator to distinguish normal from abnormal samples.

\subsubsection{MCMC}
\label{sec:__mcmc}

As the foundation for subsequent modules, MCMC introduces key multi-class contextual centers.
Without MCMC, model variant (B) uses a single sample as the center instead of the averaged center.
Since a single sample fails to capture contextual information effectively,
variant (B) shows a significant performance drop across all three datasets compared to our complete method,
with the largest decline on MPDD due to its highly dispersed part distribution.
Additionally, an ablation experiment reveals that removing the feature adapter ${{A}_{\theta}}$ leads to a -2.6\% drop in I-AUROC on MVTec AD compared to the complete CRAS,
demonstrating the importance of domain shift mitigation.

\begin{figure}[t]
  \centering
  \includegraphics[width=0.95\linewidth]{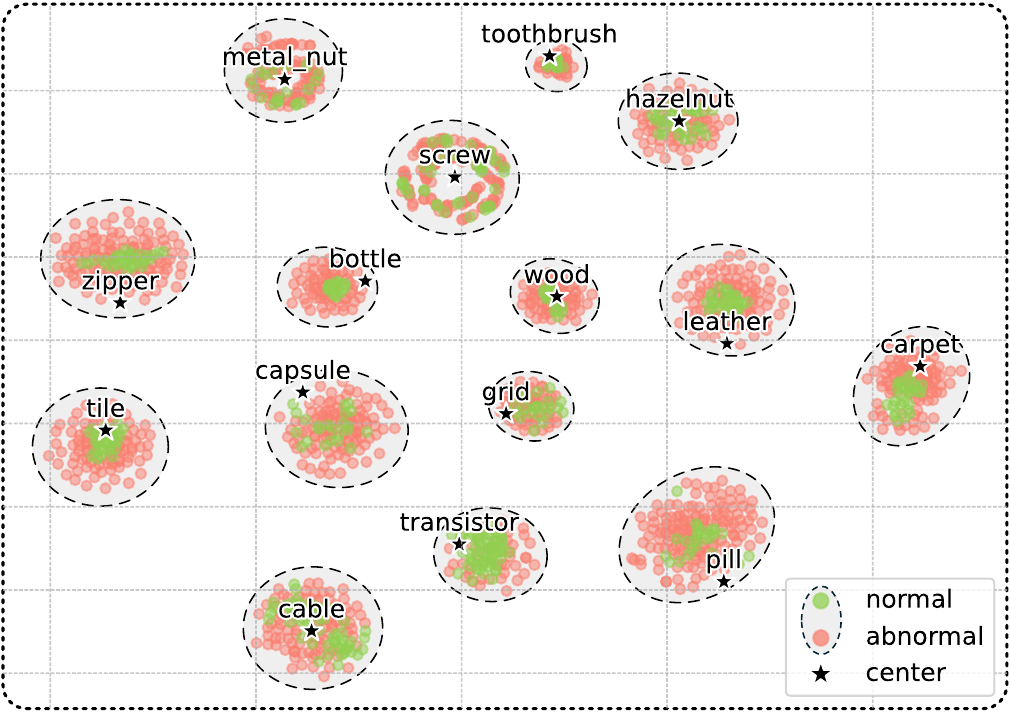}
  \caption{t-SNE visualization of adapted sample-level feature distributions on the MVTec AD dataset.
  All normal and abnormal samples are accurately matched to their category-specific global centers.}
  \label{fig:tsne_center}
\end{figure}

\subsubsection{HPI}
\label{sec:__HPI}

HPI is designed to generate more accurate recomposed centers at the local level while ensuring inference speed.
The omission of HPI in model variant (C) implies imprecise spatial alignment of global centers, resulting in a noticeable performance drop across all datasets.
As mentioned in Section~\ref{sec:__local},
our global-to-local alignment mechanism not only improves I-AUROC by +0.4\% on MVTec AD but also reduces computational complexity,
achieving a 5$\times$ speed-up in CPU inference compared to direct local matching with multi-class centers,
while maintaining comparable efficiency on GPU due to its parallel computation capability.
To validate the accuracy of category identification, we conducted additional experiments on all 15 categories of the MVTec AD dataset.
All test samples were correctly matched to their corresponding class centers, confirming the effectiveness of the proposed matching strategy.
This is further supported by the t-SNE visualization of adapted sample-level feature distributions shown in Fig.~\ref{fig:tsne_center},
where all normal and abnormal samples are distributed close to their centers.
Although misidentification may still occur in rare cases where categories share highly similar morphological patterns,
the local patch alignment step helps to refine spatial correspondence and reduce the impact on anomaly detection performance.

\subsubsection{DAFS}
\label{sec:__dafs}

DAFS generates critical anomaly features for normal features.
Model variant (D) without DAFS only utilizes fixed-variance Gaussian noise,
making synthetic anomalies more prone to overlap with normal patterns in each category.
This results in a significant performance drop on both image-level and pixel-level metrics, with the largest decline of -2.5\% in P-AUROC on the MPDD dataset.
As shown in the left table of Table~\ref{tab:dafs_crd},
we observe that DAFS performs best when the distance ratio is proportional to
the Gaussian norm $g$ and inversely proportional to the residual norm $r$ in Eq.~\ref{eq:dist_ratio}.
Here, $\propto$ denotes proportionality.
Since removing scaling increases the variability in synthetic anomaly distributions,
the performance drops by -0.4\% in both I-AUROC and P-AUROC.
Moreover, setting $\beta$ to 0 results in incorrect ratio relationships,
which hinder the model's ability to capture dispersed anomalous patterns,
further highlighting the effectiveness of combining both norms for optimal anomaly detection.

\subsubsection{CRD}
\label{sec:__crd}

To further enhance binary classification, CRD is proposed to learn the center residual distribution of normal and anomalous features.
The exclusion of CRD in model variant (E) fails to distinguish anomalies within gaps of multi-class raw features, leading to missed detections.
In contrast, our method CRAS achieves the most significant improvement through the integration of CRD,
with gains of +2.5\% in I-AUROC and +1.1\% in P-AUROC on the MVTec AD dataset.
As shown in the right table of Table~\ref{tab:dafs_crd},
we observe that concatenating both raw and residual features significantly improves CRD performance.
While discriminators with more parameters (\textit{w/} concatenation) show some improvement over those with fewer parameters (\textit{w/o} concatenation),
the introduction of residual features offers a more notable enhancement.
Using residual features alone improves both I-AUROC and P-AUROC by +0.8\%, outperforming the concatenation of raw features.
This is due to the valuable multi-class information retained in raw features.
Finally, the complete CRD further improves I-AUROC and P-AUROC by +0.2\% over the residual feature concatenation.

\begin{table}[t]\footnotesize
  \centering
  \caption{Performance ablation of distance ratios in DAFS and feature combinations in CRD on MVTec AD,
  as measured by I-AUROC\% and P-AUROC\%.
  The best results are highlighted in bold.}
  
    \begin{tabularx}{0.24\textwidth}{>{\hsize=60pt}X|*{2}{>{\centering\arraybackslash}X}}
      \noalign{\hrule height 0.4mm}
      \textbf{DAFS}  & I~/~P-AUROC \\
      \hline
      $\beta=0$ & 97.6~/~97.6 \\
      \textit{w/o} scaling & 97.9~/~97.6 \\
      $\propto 1/g,~\propto 1/r$ & 98.0~/~97.8 \\
      $\propto 1/r$ & 98.2~/~97.9 \\
      \hline
      \rowcolor{pink!30} $\propto g,~\propto 1/r$ & \textbf{98.3}~/~\textbf{98.0} \\
      \noalign{\hrule height 0.4mm}
    \end{tabularx}%
    \hspace{0cm}
    \begin{tabularx}{0.24\textwidth}{>{\hsize=60pt}X|*{2}{>{\centering\arraybackslash}X}}
      \noalign{\hrule height 0.4mm}
    \textbf{CRD}   & I~/~P-AUROC \\
    \hline
    raw & 95.8~/~96.9 \\
    residual & 97.7~/~97.5 \\
    raw+raw & 96.9~/~96.7 \\
    residual+residual & 98.1~/~97.8 \\
    \hline
    \rowcolor{pink!30} raw+residual & \textbf{98.3}~/~\textbf{98.0} \\
    \noalign{\hrule height 0.4mm}
    \end{tabularx}%
  \label{tab:dafs_crd}%
\end{table}%

\begin{figure*}[t]
  \centering
  \includegraphics[width=0.99\linewidth]{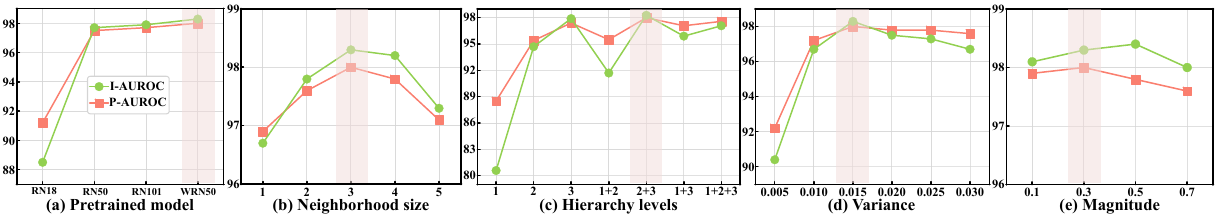}
  \caption{Quantitative results of various parameters on MVTec AD measured by I-AUROC\% and P-AUROC\%.
  (a) Dependence on pretrained model.
  (b) Selection of neighborhood size.
  (c) Concatenation of hierarchy levels.
  (d) Variance of Gaussian noise.
  (e) Magnitude of variance adjustment.
  }
  \label{fig:parameter}
\end{figure*}

\subsection{Parameter Analyses}
\label{sec:_parameter}

In this section, we have conducted a series of experiments to analyze the impact of various hyperparameters on CRAS.

\subsubsection{Dependence on Pretrained Model}
\label{sec:__pretrained}

The pretrained model $\phi$ serves as the backbone of the feature extractor $E_\phi$ for the MCMC module.
As shown in Fig.~\ref{fig:parameter}(a), anomaly detection performance positively correlates with the number of parameters.
Although the commonly-used WideResNet50 achieves the best performance,
the performance differences among backbones (except for ResNet18 \cite{he2016deep} with the fewest parameters) are minimal.
Therefore, our method exhibits robustness with respect to different backbones.

\subsubsection{Selection of Neighborhood Size}
\label{sec:__neighbor}

The neighborhood size \(p\) in feature extractor \({{E}_{\phi}}\) for the MCMC module is utilized as the patch size for feature aggregation.
Fig.~\ref{fig:parameter}(b) shows that the optimal neighborhood size is \mbox{\(p=3\)}.
Smaller patch sizes degrade performance due to insufficient spatial information aggregation,
while larger patch sizes also degrade performance due to the loss of detailed local information.

\subsubsection{Concatenation of Hierarchy Levels}
\label{sec:__concat_level}

The hierarchy levels of aggregated features are concatenated to capture low-level and high-level features in the MCMC module.
As shown in Fig.~\ref{fig:parameter}(c), the concatenation of hierarchy levels 2 and 3 achieves the best performance. 
Since hierarchy level 3 contains the deepest semantic information, it contributes the most to performance improvement.

\begin{table}[t]\footnotesize
  \centering
  \caption{Robustness evaluation on MVTec AD under different industrial disturbances, as measured by the degradation in I-AUROC\%/P-AUROC\%.
  The best results are in bold.}
  \begin{tabularx}{0.49\textwidth}{p{2.75cm}|*{3}{>{\centering\arraybackslash}X}|>{\centering\arraybackslash\columncolor{pink!30}}X}
    \noalign{\hrule height 0.4mm}
    Method~$\rightarrow$ & SimpleNet & GLASS & PBAS  & \\
    Disturbance~$\downarrow$ & \cite{liu2023simplenet}   & \cite{chen2025unified}   & \cite{chen2024progressive}   &  \multirow{-2}{*}{\textbf{CRAS}} \\
    \hline
    + adversarial noise & -2.3/-4.0 & -8.6/-4.3 & -4.8/-6.1 & \textbf{-1.9/-0.5} \\
    + illumination variation & -8.0/-3.4 & -9.4/-4.2 & -9.5/-4.5 & \textbf{-6.7/-1.8} \\
    + equipment vibration & -6.5/-1.9 & -13.7/-7.0 & -18.6/-8.5 & \textbf{-6.0/-1.2} \\
    \hline
    Mean & -5.6/-3.1 & -10.6/-5.2 & -11.0/-6.4 & \textbf{-4.9/-1.2} \\
    \noalign{\hrule height 0.4mm}
    \end{tabularx}%
  \label{tab:disturb}%
\end{table}%

\subsubsection{Variance of Gaussian Noise}
\label{sec:__gaussian_var}

The initial variance of Gaussian noise $\sigma$ in the DAFS module is crucial for anomaly feature synthesis.
Fig.~\ref{fig:parameter}(d) demonstrates that the optimal variance is \mbox{\(\sigma=0.015\)}.
Smaller variances significantly reduce performance due to positive-negative overlap,
while larger variances slightly reduce performance by making them overly distinguishable.

\subsubsection{Magnitude of Variance Adjustment}
\label{sec:__synthesis_range}

In Eq.~\ref{eq:dist_ratio} of the DAFS module, the factor $\beta$ controls the magnitude of variance adjustment.
As indicated in Fig.~\ref{fig:parameter}(e),
the impact of $\beta$ on performance is relatively minor because the model has learned anomaly patterns with similar distributions.
Compared to the second-best result at \mbox{$\beta=0.5$},
the optimal result at \mbox{$\beta=0.3$} lags by -0.1\% in I-AUROC but leads by +0.2\% in P-AUROC.

\begin{figure}[t]
   \centering
   \includegraphics[width=0.99\linewidth]{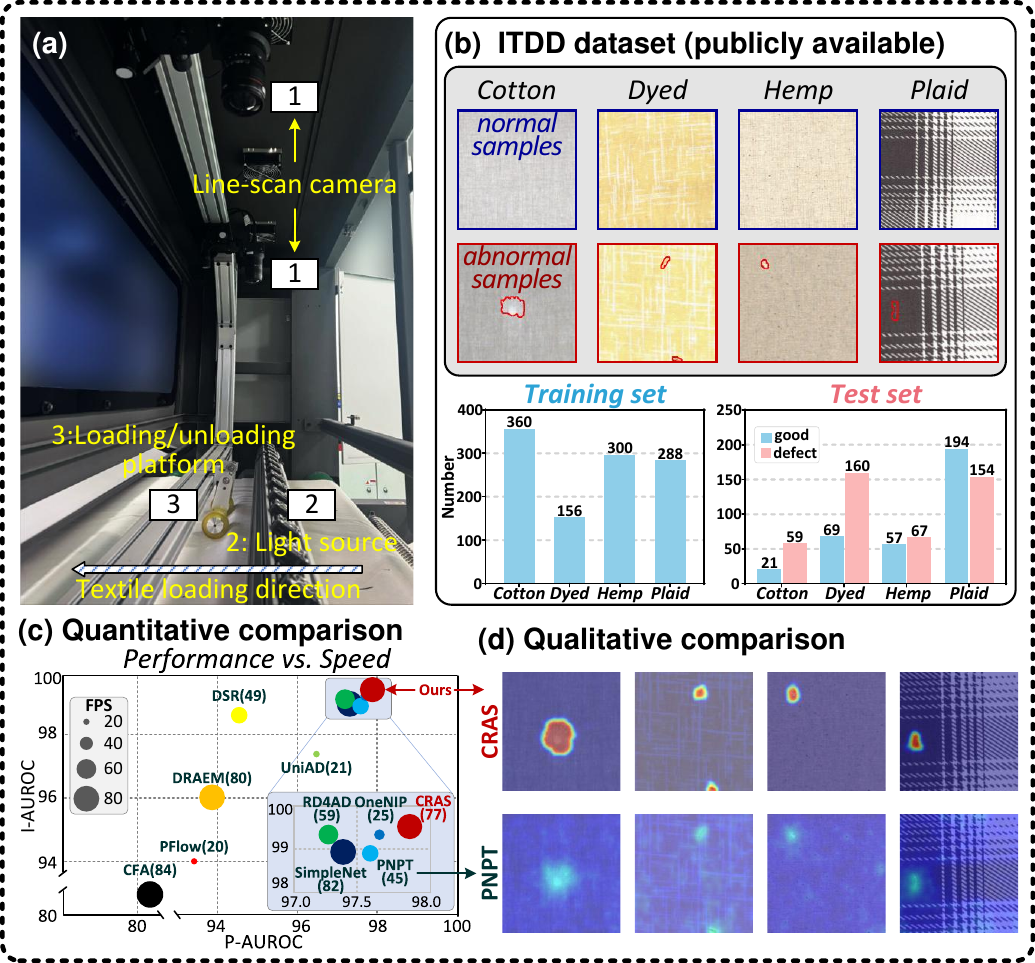}
   \caption{Experimental verification of CRAS on real-world application.
   (a) Automatic optical inspection equipment.
   (b) Industrial Textile Defect Detection (ITDD) dataset.
   (c) Quantitative results.
   (d) Qualitative results.
   }
   \label{fig:application}
\end{figure}

\subsubsection{Robustness under industrial disturbances}
\label{sec:__robustness}

To evaluate the practical robustness of CRAS,
we introduce three common types of industrial disturbances into each test image in MVTec AD.
Specifically, we simulate adversarial noise using $\mathcal{N}(0, 0.01^2)$,
illumination variation with color jitter (brightness, contrast, saturation, hue = 0.1),
and equipment vibration with random rotations up to 5 degrees.
As summarized in Table~\ref{tab:disturb}, CRAS exhibits the lowest average performance drop across all scenarios,
confirming its reliability under real-world deployment challenges.
This is attributed to our center-aware residual learning, which filters out irrelevant information.

\subsection{Real-world Industrial Application}
\label{sec:_itdd}

To further assess the generalization and efficiency of our method in industrial scenarios,
we applied it to a real-world task of detecting textile defects.
Our automatic optical inspection equipment is shown in Fig.~\ref{fig:application}(a),
which is capable of capturing high-resolution images of textiles.
We further cropped them into images of size $512 \times 512$ using a sliding window.
As shown in Fig.~\ref{fig:application}(b), 
we collected and annotated the ITDD dataset, comprising 1,885 images across four fabric types, for multi-class industrial anomaly detection.
Specifically, the training set includes 1,104 normal samples, while the test set consists of 341 normal samples and 440 abnormal samples.
The quantitative and qualitative comparisons are presented in Fig.~\ref{fig:application}(c-d), respectively.
The proposed CRAS outperforms other methods, achieving an I-AUROC of 99.4\% and a P-AUROC of 97.8\%.
It also yields fewer missed detections, cleaner background segmentation, and maintains a competitive inference speed of 77 FPS.
\section{Conclusion}
\label{sec:conclu}

In this paper, we propose the Center-aware Residual Anomaly Synthesis (CRAS) method, a unified model for multi-class industrial anomaly detection. 
CRAS addresses the challenges of inter-class interference and intra-class overlap through its innovative designs.
By leveraging the CRD module, our method unifies samples from different categories into a shared center, effectively reducing missed detections.
To address over-detection, the DAFS module adaptively adjusts noise variance, enhancing detection precision.
Additionally, the MCMC and HPI modules are introduced to maintain fast inference speed.
Comprehensive experiments across four datasets demonstrate the superior performance of CRAS in multi-class anomaly detection.

Although current vision-only models excel at detecting structural anomalies like surface scratches,
they struggle to identify semantic anomalies such as incorrectly assembled industrial components.
Therefore, VLMs are gaining growing attention for zero-shot anomaly detection tasks.
In the future, we will explore the potential of integrating CRAS with VLMs for detecting more challenging defects in general scenarios.

\bibliography{main}
\bibliographystyle{IEEEtran}

\begin{IEEEbiography}[{\includegraphics[width=0.99in,height=1.25in,clip,trim=0cm 0.2cm 0cm 0.2cm]{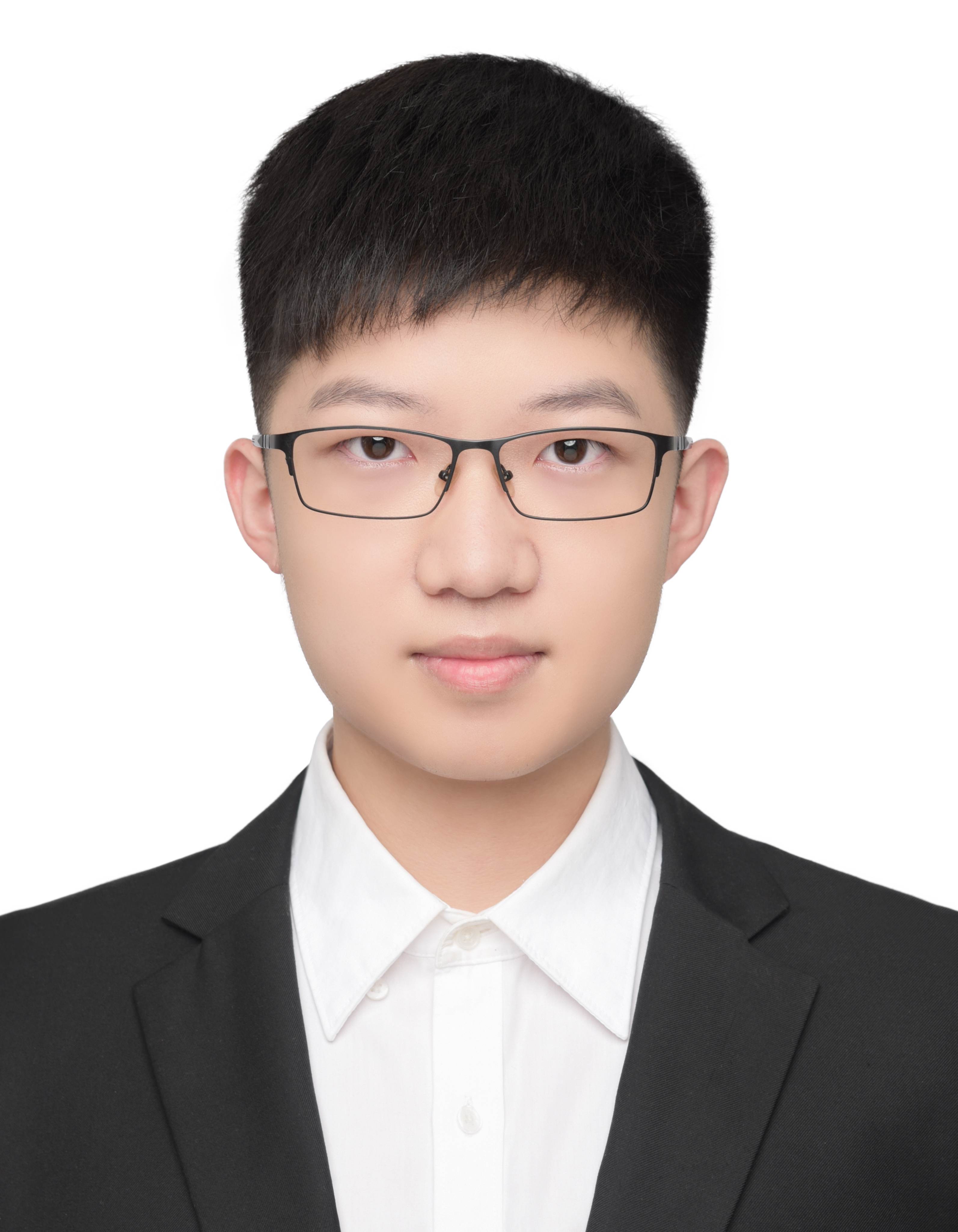}}]{Qiyu Chen}
    received the B.E. degree in Communication Engineering from Tongji University, Shanghai, China, in 2021.
    He is currently pursuing the Ph.D. degree with the Institute of Automation, Chinese Academy of Sciences (IACAS), Beijing, China,
    and also with the School of Artificial Intelligence, University of Chinese Academy of Sciences, Beijing.

    His research interests include deep learning, computer vision, and anomaly detection.
\end{IEEEbiography}

\begin{IEEEbiography}[{\includegraphics[width=0.99in,height=1.25in,clip,trim=0cm 0.5cm 0cm 0.5cm]{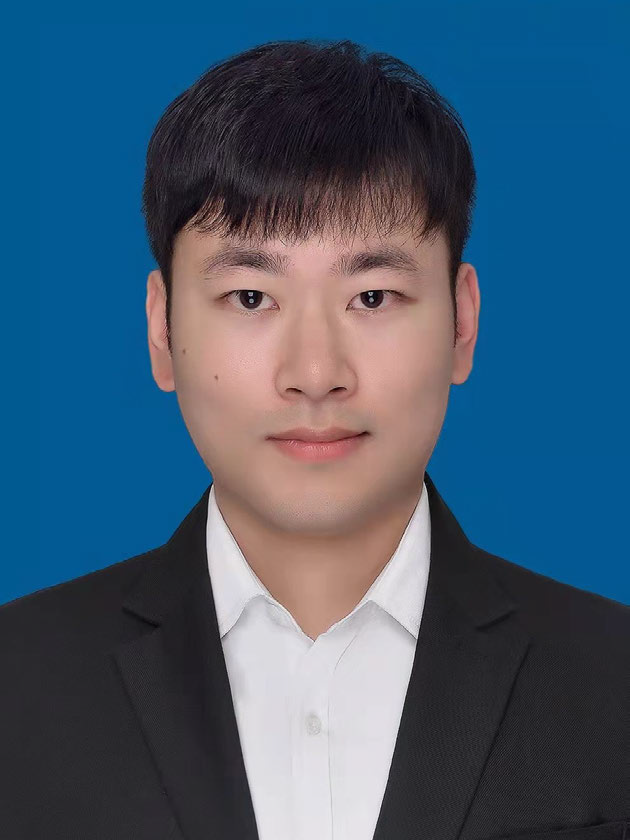}}]{Huiyuan Luo}
    received the B.S. degree from Harbin Institute of Technology, Weihai, China, in 2016, 
    and the Ph.D. degree from the Changchun Institute of Optics, Fine Mechanics and Physics, Chinese Academy of Science, Changchun, China, in 2021.
    Since 2022, he has been a postdoctor and assistant researcher at the Institute of Automation, Chinese Academy of Sciences (IACAS), Beijing.

    He has been engaged in saliency detection, industrial anomaly detection, unsupervised learning, and intelligent manufacturing.
\end{IEEEbiography}

\begin{IEEEbiography}[{\includegraphics[width=0.99in,height=1.25in,clip,trim=0cm 0.9cm 0cm 0.9cm]{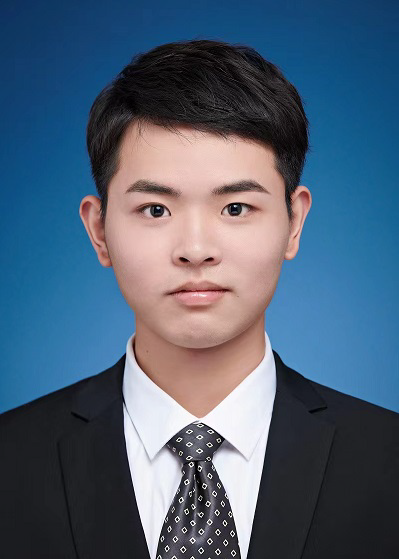}}]{Haiming Yao}
    (Graduate Student Member, IEEE)
    received a B.S. degree (Hons.) from the School of Mechanical Science and Engineering, Huazhong University of Science and Technology, Wuhan, China, in 2022.
    He is pursuing a Ph.D. degree with the Department of Precision Instrument, Tsinghua University, Beijing, China.

    His research interests include visual anomaly detection, deep learning, visual understanding, and artificial intelligence for science.
\end{IEEEbiography}

\begin{IEEEbiography}[{\includegraphics[width=0.99in,height=1.25in,clip,trim=0cm 0cm 0cm 0cm]{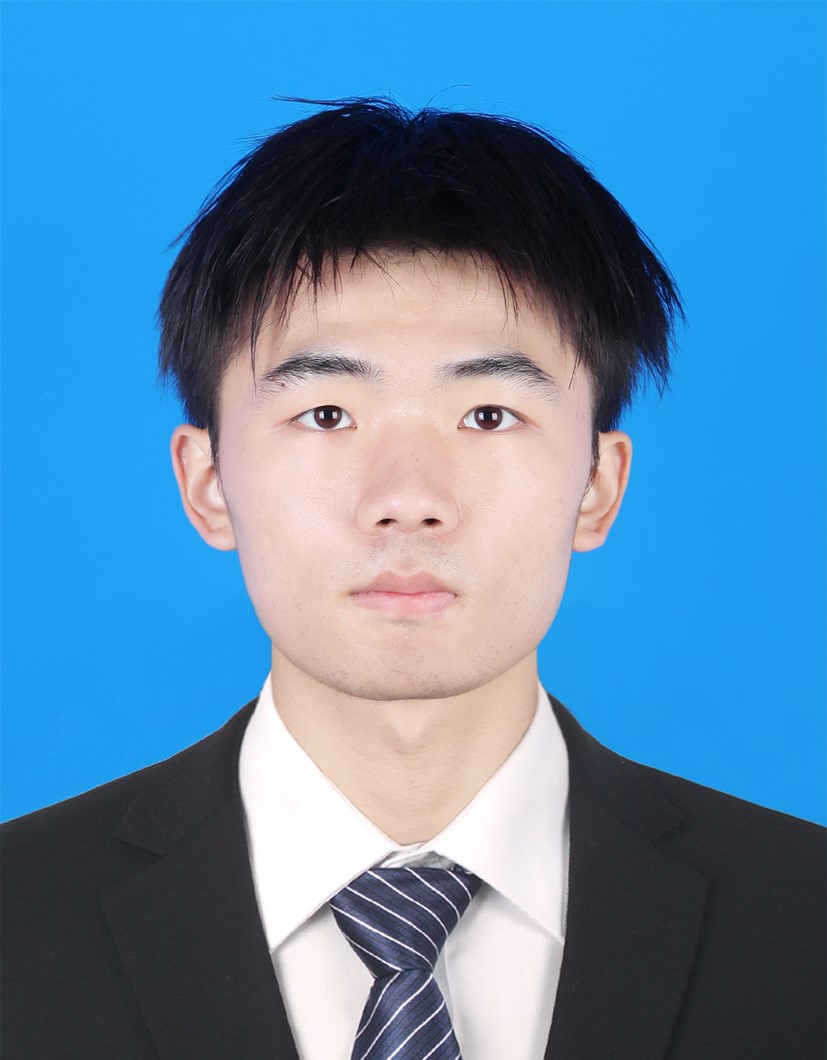}}]{Wei Luo}
    (Student Member, IEEE)
    received a B.S. degree from the School of Mechanical Science and Engineering, Huazhong University of Science and Technology, Wuhan, China, in 2023.
    He is pursuing a Ph.D. degree with the Department of Precision Instrument, Tsinghua University, Beijing, China.
    
    His research interests include deep learning, anomaly detection, and machine vision.
\end{IEEEbiography}

\begin{IEEEbiography}[{\includegraphics[width=0.99in,height=1.25in,clip,trim=0cm 0.3cm 0cm 0.3cm]{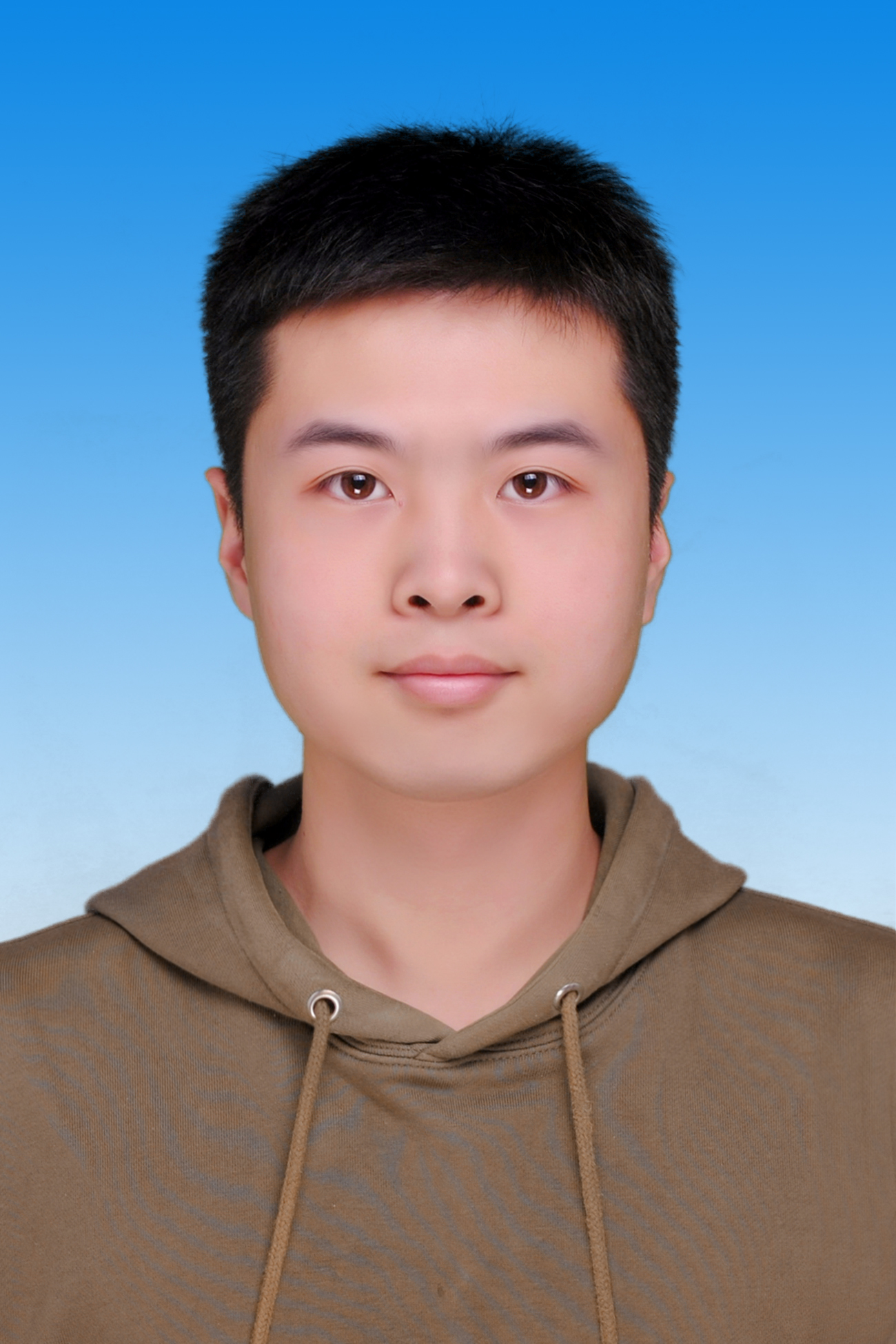}}]{Zhen Qu}
    received the B.Sc. degree from Xidian University, Xi'an, China, in 2022.
    He is currently working towards the Ph.D. degree with the Institute of Automation, Chinese Academy of Sciences (IACAS), Beijing, China,
    and also with the School of Artificial Intelligence, University of Chinese Academy of Sciences, Beijing.
    
    His research interests include machine learning and visual inspection.
\end{IEEEbiography}

\begin{IEEEbiography}[{\includegraphics[width=0.99in,height=1.25in,clip,trim=0cm 0cm 0cm 0cm]{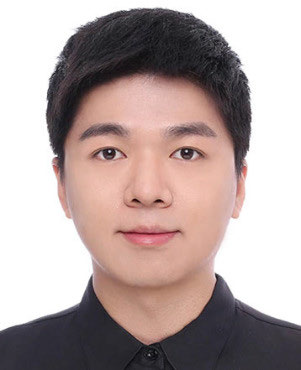}}]{Chengkan Lv}
    (Member, IEEE)
    received the B.S. degree from Shandong University, Jinan, China, in 2017, and the Ph.D. degree
    from the Institute of Automation, Chinese Academy of Sciences, Beijing, China, in 2022.

    His research interests include neural networks, computer vision, and anomaly detection.
\end{IEEEbiography}

\begin{IEEEbiography}[{\includegraphics[width=0.99in,height=1.25in,clip,trim=0cm 0.4cm 0cm 0.4cm]{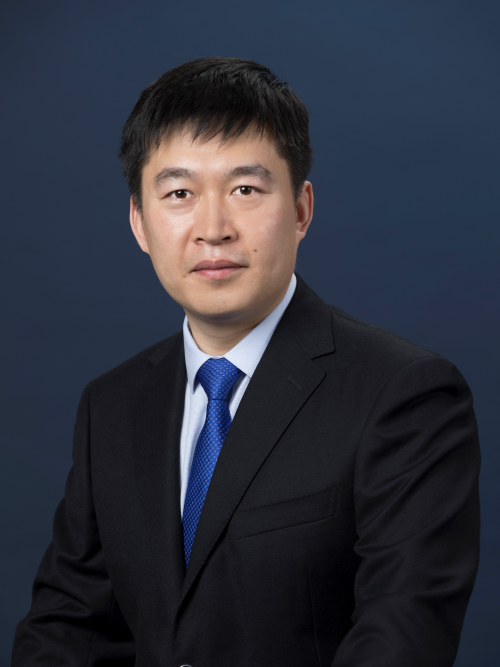}}]{Zhengtao Zhang}
    (Member, IEEE)
    received the B.S. degree from the China University of Petroleum, Dongying, China, in 2004, the M.S. degree
    from the Beijing Institute of Technology, Beijing, China, in 2007, and the Ph.D. degree
    in control science and engineering from the Institute of Automation, Chinese Academy of Sciences (IACAS), Beijing, in 2010.
    
    He is currently a Professor at IACAS and a Doctoral Supervisor at the School of Artificial Intelligence, University of Chinese Academy of Sciences.
    He also holds a part-time position as the Director at the Binzhou Institute of Technology, Binzhou, Shandong, China.
    His research interests include visual measurement, microassembly, and automation.
\end{IEEEbiography}


\end{document}